\documentclass[10pt,twocolumn,letterpaper]{article}

\usepackage{iccv}
\usepackage{times}
\usepackage{epsfig}
\usepackage{graphicx}
\usepackage{amsmath}
\usepackage{amssymb}

\usepackage{color}
\usepackage{xcolor}
\usepackage{booktabs}
\usepackage{algorithm}
\usepackage{algorithmic}
\usepackage{multirow}

\usepackage{stfloats}
\usepackage{bbding}
\usepackage{subfigure}

\usepackage[pagebackref=true,breaklinks=true,letterpaper=true,colorlinks,bookmarks=false]{hyperref}

\usepackage[capitalize]{cleveref}
\crefname{section}{Sec.}{Secs.}
\Crefname{section}{Section}{Sections}
\Crefname{table}{Table}{Tables}
\crefname{table}{Tab.}{Tabs.}

 \iccvfinalcopy 

\def\iccvPaperID{8856} 

\ificcvfinal\fi

\begin{document}

\title{Towards Viewpoint-Invariant Visual Recognition via Adversarial Training}


\author{
Shouwei Ruan$^{1}$, Yinpeng Dong$^{2,3}$, Hang Su$^{2,4,5*}$, Jianteng Peng$^{6}$, Ning Chen$^{2}$, Xingxing Wei$^{1}$\thanks{Corresponding author.} \\
$^{1}$ Institute of Artificial Intelligence, Beihang University, Beijing 100191, China \\
  $^{2}$ Dept. of Comp. Sci. and Tech., Institute for AI, Tsinghua-Bosch Joint ML Center,\\
  THBI Lab, BNRist Center, Tsinghua University, Beijing 100084, China \\
  $^{3}$ RealAI \hspace{0.5ex} $^{4}$ Peng Cheng Laboratory \hspace{0.5ex} $^{5}$ Pazhou Laboratory (Huangpu), Guangzhou, China  \hspace{0.5ex} $^{6}$ OPPO \\ 
  \scriptsize{\texttt{\{dongyinpeng,suhangss,ningchen\}@tsinghua.edu.cn, \{shouweiruan,xxwei\}@buaa.edu.cn,~pengjianteng@oppo.com}}
}

\maketitle
\ificcvfinal\thispagestyle{empty}\fi

\begin{abstract}
\vspace{-0.2cm}
Visual recognition models are not invariant to viewpoint changes in the 3D world, as different viewing directions can dramatically affect the predictions given the same object. Although many efforts have been devoted to making neural networks invariant to 2D image translations and rotations, viewpoint invariance 
  is rarely investigated. As most models process images in the perspective view, it is challenging to impose invariance to 3D viewpoint changes based only on 2D inputs. 
  Motivated by the success of adversarial training in promoting model robustness, we propose Viewpoint-Invariant Adversarial Training (VIAT) to improve viewpoint robustness of common image classifiers. 
   By regarding viewpoint transformation as an attack, VIAT is formulated as a minimax optimization problem, where the inner maximization characterizes diverse adversarial viewpoints by learning a Gaussian mixture distribution based on a new attack GMVFool, while the outer minimization trains a viewpoint-invariant classifier by minimizing the expected loss over the worst-case adversarial viewpoint distributions. 
  To further improve the generalization performance, a distribution sharing strategy is introduced leveraging the transferability of adversarial viewpoints across objects. Experiments validate the effectiveness of VIAT in improving the viewpoint robustness of various image classifiers based on the diversity of adversarial viewpoints generated by GMVFool. 
  
\end{abstract}

\vspace{-0.6cm}
\section{Introduction}
\label{sec:intro}
\vspace{-0.1cm}

\begin{figure}[t]
  \centering
  \includegraphics[width=0.96\linewidth]{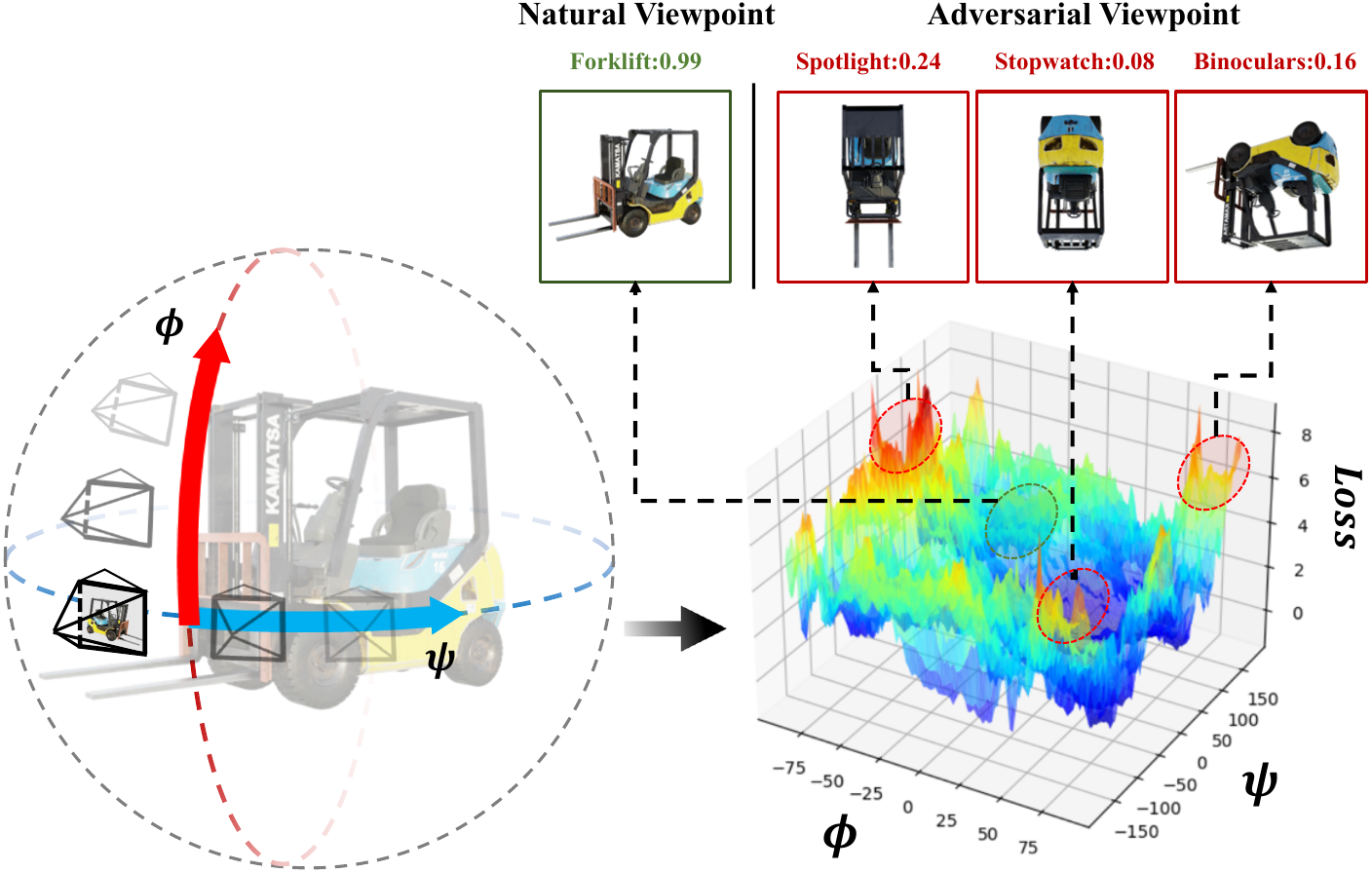}
   \caption{An illustration of viewpoint changes on model performance. We show the loss landscape w.r.t. yaw and pitch of the camera, which demonstrates multiple regions of adversarial viewpoints (We use ResNet-50 as the target model~\cite{he2016deep}). 
   }
   \label{fig:0} \vspace{-0.3cm}
\end{figure}

\begin{figure*}
  \centering
  \includegraphics[width=0.97\linewidth]{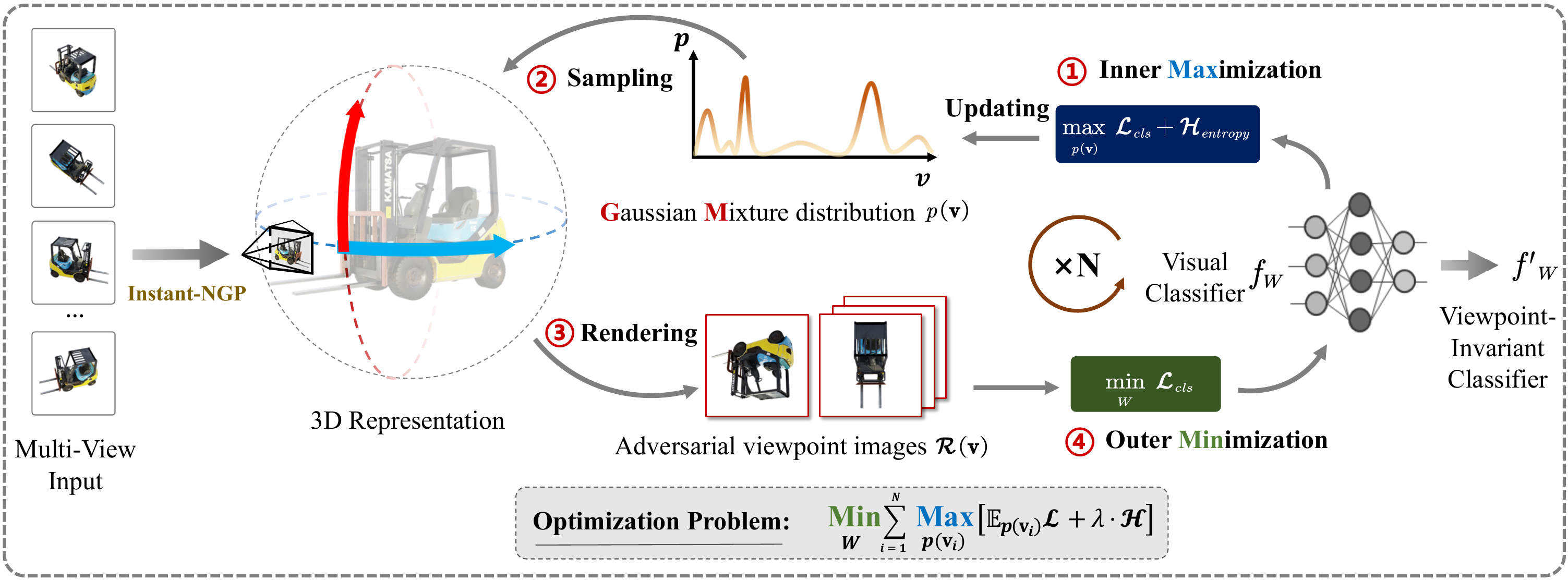}
   \caption{An overview of our VIAT framework. We first train the NeRF representation of each object given multi-view images. The inner maximization learns a Gaussian mixture distribution of adversarial viewpoints by maximizing the expectation of classification loss and entropy regularization. The outer minimization samples adversarial viewpoints from the optimized distributions and renders 2D images from adversarial viewpoints, which are fed into the network along with clean samples to train viewpoint-invariant classifiers.} \vspace{-0.5cm}
   \label{fig:framework}
\end{figure*}

The ability of learning invariant representations is highly desirable in numerous computer vision tasks~\cite{bengio2013representation,goodfellow2009measuring} and is conducive to model robustness under semantic-preserving image transformations. Previous works~\cite{engstrom2019exploring,zhang2019making,cohen2016group,sifre2013rotation} have striven to make visual recognition models invariant to image translation, rotation, reflection, and scaling. However, they mainly consider invariances to 2D image transformations, leaving the \emph{viewpoint} transformation~\cite{zemel1990discovering} in the 3D world less explored. It has been shown that visual recognition models are susceptible to viewpoint changes~\cite{alcorn2019strike,barbu2019objectnet,dong2022viewfool}, exhibiting a significant gap from the human vision that can robustly recognize objects under different viewpoints~\cite{biederman1987recognition}. Due to the naturalness and prevalence of viewpoint variations in safety-critical applications (\eg, autonomous driving, robotics, surveillance, \etc), it is thus imperative to endow visual recognition models with viewpoint invariance.

Despite the importance, it is extremely challenging to build viewpoint-invariant visual recognition models since typical networks take 2D images as inputs without inferring the structure of 3D objects. As an effective data-driven approach, adversarial training augments training data with adversarially generated samples under a specific threat model and shows promise to improve model invariance/robustness to additive adversarial perturbations~\cite{madry2017towards,zhang2019theoretically,wei2022physically}, 
image translation and rotation~\cite{engstrom2019exploring}, geometric transformations \cite{kanbak2018geometric}, \etc. 
However, it is non-trivial to directly apply adversarial training to improving viewpoint robustness due to the difficulty of generating the worst-case adversarial viewpoints. 
A pioneering work~\cite{dong2022viewfool} proposes \emph{ViewFool}, which encodes real-world 3D objects as Neural Radiance Fields (NeRF)~\cite{mildenhall2020nerf} given multi-view images and performs black-box optimization for generating a distribution of adversarial viewpoints.
Though effective, ViewFool only adopts a unimodal Gaussian 
distribution, which is inadequate to characterize multiple local maxima of the loss landscape w.r.t. viewpoint changes, as shown in Fig.~\ref{fig:0}. 
We verify that this can lead to overfitting of adversarial training to the specific attack. Besides, ViewFool is time-consuming to optimize, making adversarial training intractable. 

To address these problems, in this paper, we propose \textbf{ Viewpoint-Invariant Adversarial Training (VIAT)}, the first framework to improve the viewpoint robustness of visual recognition models via adversarial training. As shown in Fig.~\ref{fig:framework}, VIAT is formulated as a distribution-based minimax problem, in which the inner maximization aims to optimize the distribution of diverse adversarial viewpoints while the outer minimization aims to train a viewpoint-invariant classifier by minimizing the expected loss over the worst-case adversarial viewpoint distributions. 
To address the limitations of ViewFool, we propose \textbf{GMVFool} as a practical solution to the inner problem, which generates a Gaussian mixture distribution of adversarial viewpoints for each object, with increased diversity to mitigate overfitting of adversarial training. 
To accelerate training, we adopt a stochastic optimization strategy to reduce the time cost of training and adopt Instant-NGP~\cite{mueller2022instant}, a fast variant of NeRF, to improve the efficiency of the optimizing process. 
In outer maximization, to further improve generalization, we propose a distribution-sharing strategy given the observation that adversarial viewpoint distributions are transferable across objects within the same class. We fine-tune classifiers on a mixture of natural and sampled adversarial viewpoint images to improve their viewpoint invariance.

To verify VIAT's ability of training a viewpoint-invariant model,  a multi-view dataset is required. However, previous datasets~\cite{geusebroek2005amsterdam,kanezaki2018rotationnet,reizenstein2021common,collins2022abo} usually have limited realism and viewpoint range, posing challenges when applying them to this topic. To address this, we devoted significant effort to creating a new multi-view dataset---\textbf{IM3D}, which contains 1k typical synthetic 3D objects from 100 classes, tailored specifically for ImageNet categories. IM3D has several notable advantages compared to previous datasets, as shown in Table~\ref{table: dataset comparison}: (1) It covers more categories. (2) It utilizes physics-based rendering (PBR) technology\footnote{\scriptsize A 3D modeling and rendering technique that enables physically realistic effects.} to produce realistic images. (3) It has accurate camera pose annotations and is sampled from a spherical space, leading to better reconstruction quality and exploration of the entire 3D space. Thus, we mainly use it for training and further evaluating our method on other multi-view datasets. We will release our IM3D dataset, which includes multi-view images, 3D source files, and corresponding Instant-NGP weights.

We conduct extensive experiments to validate the effectiveness of both GMVFool and VIAT for generating adversarial viewpoints and improving the viewpoint robustness of image classifiers. Experimental results show that GMVFool characterizes more diverse adversarial viewpoints while maintaining high attack success rates. Based on it, VIAT significantly improves the viewpoint robustness of image classifiers ranging from ResNet \cite{he2016deep} to Vision Transformer (ViT)~\cite{dosovitskiy2020image} and shows superior performance compared with alternative baselines. 
Moreover, we construct a new out-of-distribution (OOD) benchmark---\textbf{ImageNet-V+}, containing nearly 100k images from the adversarial viewpoints found by GMVFool. It is $10\times$ larger than the previous ImageNet-V benchmark~\cite{dong2022viewfool}. We hope to serve it as a standard benchmark for evaluating viewpoint robustness in the future.

\begin{table}[htb]
\scriptsize
\setlength\tabcolsep{5.5pt}
\renewcommand\arraystretch{1.0}
\centering

\begin{tabular}{l|c|c|c|c|c}
\hline
Dataset           & \#Objects & \#Classes & PBR  & Full 3D  & Spherical Pose \\ \hline \hline 
ALOI \cite{geusebroek2005amsterdam}              & 1K        & -        & \XSolidBrush   &\XSolidBrush & \Checkmark     \\
MIRO \cite{kanezaki2018rotationnet}         & 120       & 12        & \XSolidBrush  & \XSolidBrush & \Checkmark         \\
OOWL \cite{ho2019catastrophic}        & 500       & 25        &  \XSolidBrush   & \XSolidBrush & \XSolidBrush          \\
CO3D \cite{reizenstein2021common}              & 18.6K     & 50        &  \XSolidBrush  &  \XSolidBrush   & \XSolidBrush          \\
ABO \cite{collins2022abo}               & 8K        & 63        &  \Checkmark &  \Checkmark   & \XSolidBrush          \\
Dong \etal \cite{dong2022viewfool}               & 100        & 85        &  \Checkmark &  \Checkmark   & \Checkmark \\
\textbf{IM3D (Ours)} & 1K        & 100       &  \Checkmark  &  \Checkmark   & \Checkmark        \\ \hline
\end{tabular}
\vspace{-0.1cm}
\caption{Comparison of our multi-view dataset with others.}
\label{table: dataset comparison}
\vspace{-0.3cm}
\end{table}






\section{Related Work}
\label{sec:formatting}

\subsection{Robustness to Viewpoint Transformation}
\vspace{-0.5ex}
Since deep learning models have been applied in numerous safety-critical fields, it is necessary to study the robustness of visual recognition models to 3D viewpoint transformations. The ObjectNet \cite{barbu2019objectnet}, OOD-CV \cite{zhao2022ood}, and ImageNet-R \cite{hendrycks2021many} datasets introduce images including various uncommon camera viewpoints, object poses, and object shapes to evaluate out-of-distribution (OOD) generalization under viewpoint changes. But they are unable to evaluate the performance under the worst-case viewpoint transformation. Alcorn \etal \cite{alcorn2019strike} generate adversarial perspective samples for 3D objects using a differentiable renderer and find that the model is highly susceptible to viewpoint transformation. Hamdi \etal \cite{hamdi2020towards} demonstrate the effect of viewpoint perturbation on the model performance of 3D objects and use integral boundary optimization to find robust viewpoint regions for the model. Madan \etal \cite{madan2020and} introduces diverse category-viewpoint combination images through digital objects and scenes to improve the model's generalization to OOD viewpoints. However, these methods all require 3D models. Dong \etal \cite{dong2022viewfool} further proposes ViewFool, which uses NeRF to build 3D representations of objects within multi-view images and optimizes the adversarial viewpoint distribution under an entropy regularizer. But it lacks the ability to discover diverse adversarial viewpoints. Our work differs from them mainly in that we focus on improving the viewpoint robustness of models rather than attacking them and then design a more efficient method to generate diverse adversarial viewpoints for this purpose.

\vspace{-0.3ex}
\subsection{Adversarial Training}\vspace{-0.3ex}
The concept of adversarial training (AT) is introduced by Goodfellow \etal~\cite{goodfellow2014explaining} and is widely recognized as the most effective way to enhance the robustness of deep learning models~\cite{athalye2018robustness,bai2021recent}. Based on the classical AT frameworks such as PGD-AT~\cite{madry2017towards}, previous studies have proposed improvement strategies from different aspects~\cite{tramer2017ensemble,shafahi2019adversarial,zhu2019freelb,zhang2019theoretically,pang2020bag,jia2023improving}. Adversarial training is being widely adopted for various deep learning tasks, such as visual recognition~\cite{goodfellow2014explaining, madry2017towards,zhang2019theoretically,zhao2022enhanced}, point cloud recognition~\cite{liu2019extending, zhao2020isometry,wang2022art}, and text classification~\cite{miyato2016adversarial, pan2022improved}. For viewpoint robustness, Alcorn \etal~\cite{alcorn2019strike} demonstrate that adversarial training can have an effect. They generate adversarial viewpoint images by the renderer. However, it only improves the robustness of known objects, while the generalization for unseen could be better. The difference from our work is that we don't rely on traditional renderers and 3D information of objects and can significantly improve the model's adversarial viewpoint generalization ability for unseen objects.


\vspace{-0.3ex}
\section{Methodology}
\vspace{-0.3ex}
The proposed Viewpoint-Invariant Adversarial Training (VIAT) is given here. We first introduce the background of NeRF in Sec.~\ref{sec:nerf} and the problem formulation in Sec.~\ref{sec:Formulation}, and then present the solutions of VIAT to the inner maximization in Sec.~\ref{sec:Attack} and outer minimization in Sec.~\ref{sec:Defense}. An overview of VIAT is shown in Fig.~\ref{fig:framework}.

\subsection{Preliminary on Neural Radiance Fields (NeRF)} \vspace{-0.3ex}
\label{sec:nerf}


Given a set of multi-view images, NeRF~\cite{mildenhall2020nerf} has the ability to implicitly represent the object/scene as a continuous volumetric radiance field $F:(\mathbf{x},\mathbf{d})\rightarrow (\mathbf{c}, \tau)$, 
where $F$ maps the 3D location $ \mathbf{x}\in \mathbb{R} ^3 $ and the viewing direction $ \mathbf{d}\in \mathbb{S} ^2 $ to an emitted color $ \mathbf{c}\in [0,1]^3 $ and a volume density $\tau\in\mathbb{R}^+ $. Then, using the volume rendering with stratified sampling, we can render a 2D image from a specific viewpoint. Given a camera ray $\mathbf{r}(t) = \mathbf{o} + t\mathbf{d} $ emitted from the camera center $\mathbf{o}$ through a pixel on the image plane, the expected color $\hat{C}(\mathbf{r})$ of the pixel can be calculated by a discrete set of sampling points $\{t_m\}_{m=1}^{M}$ as\vspace{-0.5ex}
\begin{equation} \hat{C}(\mathbf{r})  = \sum_{m=1}^M T(t_m)\cdot \alpha (\tau(t_m) \cdot \delta_m) \cdot \mathbf{c}(t_m),
    \label{eq:nerf_2}
\end{equation}
where $T(t_m)  = \exp(-\sum_{j=1}^{m-1}\tau(t_j)\cdot\delta_j)$,  $\tau(t_m)$ and $\mathbf{c}(t_m)$ denote the volume density and color at point $ \mathbf{r}(t_m)$, $\delta_m = t_{m+1}-t_m$ is the distance between adjacent points, and $ \alpha{(x)}=1-\exp{(-x)} $. $F$ is approximated by a multi-layer perceptron (MLP) network and optimized by minimizing the $L_2$ loss between the rendered and ground-truth pixels.

Although NeRF can render photorealistic novel views, both training and rendering are extremely time-consuming. Instant-NGP~\cite{mueller2022instant} proposes a fast implementation of NeRF by adaptive and efficient multi-resolution hash encoding. Therefore, in this paper, we adopt Instant-NGP to accelerate the training and volumetric rendering of NeRF.

\subsection{Problem Formulation} \label{sec:Formulation}

In visual recognition, viewpoint invariance indicates that a model $f(\cdot)$ can make an identical prediction given two views of the same object as follows:
\begin{equation}
f(I(\mathbf{v}_1))=f(I(\mathbf{v}_2)), \;\;\; \forall (\mathbf{v}_1, \mathbf{v}_2)
\end{equation}
where $I(\mathbf{v}_1)$ and $I(\mathbf{v}_2)$ are two images taken from arbitrary viewpoints $\mathbf{v}_1$ and $\mathbf{v}_2$ of the object. However, recent studies~\cite{barbu2019objectnet,dong2022viewfool,alcorn2019strike} have revealed that typical image classifiers are susceptible to viewpoint changes. As viewpoint variations in the 3D space cannot be simply simulated by 2D image transformations, it remains challenging to improve viewpoint invariance/robustness. Motivated by the success of adversarial training in improving model robustness, we propose \textbf{Viewpoint-Invariant Adversarial Training (VIAT)} by learning on worst-case adversarial viewpoints.

Formally, viewpoint changes can be described as rotation and translation of the camera in the 3D space~\cite{dong2022viewfool}. We let $\mathbf{v} = [\mathbf{R}, \mathbf{T}] \in \mathbb{R}^6$ denote the viewpoint parameters bounded in $[\mathbf{v}_{\min}, \mathbf{v}_{\max}]$, where $ \mathbf{R}=[\psi,\theta,\phi]$ is the camera rotation along the z-y-x axes using the Tait-Bryan angles, and $ \mathbf{T}=[\Delta _x,\Delta _y,\Delta_z]$ is the camera translation along the three axes.
Given a dataset $\{\mathrm{obj}_i\}_{i=1}^ N$ of $N$ objects and the corresponding ground-truth labels $\{y_i\}_{i=1}^N$ with $y_i\in\{1,...,C\}$, we suppose that a set of multi-view images is available for each object. With these images, we first train a NeRF model for each object using Instant-NGP to obtain a neural renderer that can synthesize new images from any viewpoint of the object. Rather than finding an adversarial viewpoint $\mathbf{v}_i$ for each object $\mathrm{obj}_i$, VIAT characterizes diverse adversarial viewpoints by learning the underlying distribution $p(\mathbf{v}_i)$, which can be formulated as a distribution-based minimax optimization problem:
\begin{equation}
\min_{\mathbf{W}}\!\sum_{i=1}^{N}   \max_{p(\mathbf{v}_i ) }\left [ \mathbb{E}_{p(\mathbf{v}_i )}\! \left [  \mathcal{L}\left(f_{\mathbf{W}}\left (\mathcal{R}(\mathbf{v}_i )  \right ), y_i\right )\right ]\! + \!\lambda\! \cdot\!\mathcal{H}(p(\mathbf{v}_i))\right ]\!,
\label{eq: AT formulation}
\end{equation}
where $\mathbf{W}$ denotes the parameters of the classifier $f_{\mathbf{W}}$, $\mathcal{R}(\mathbf{v}_i)$ is the rendered image of the $i$-th object given the viewpoint $\mathbf{v}_i$, $\mathcal{L}$ is a classification loss (\eg, cross-entropy loss), and $\mathcal{H}(p(\mathbf{v}_i))=-\mathbb{E}_{p(\mathbf{v}_i)}[\log p(\mathbf{v}_i)]$ is the entropy of the distribution $p(\mathbf{v}_i)$ to avoid the degeneration problem and help to capture more diverse adversarial viewpoints~\cite{dong2022viewfool}.

As can be seen in Eq.~\eqref{eq: AT formulation}, the inner maximization aims to learn a distribution of adversarial viewpoints under an entropic regularizer, while the outer minimization aims to optimize model parameters by minimizing the expected loss over the worst-case adversarial viewpoint distributions. The motivation of using a distribution instead of a single adversarial viewpoint for adversarial training is two-fold. First, learning a distribution of adversarial viewpoints can effectively mitigate the reality gap between the real objects and their neural representations~\cite{dong2022viewfool}. Second,
the distribution is able to cover a variety of adversarial viewpoints to alleviate potential overfitting of adversarial training, leading to better generalization performance. 

To solve the minimax problem, a general algorithm is to first solve the inner problem and then perform gradient descent for the outer problem at the inner solution in a sequential manner based on the Danskin's theorem~\cite{danskin2012theory}. Next, we introduce the detailed solutions to the inner maximization and outer minimization problems, respectively.

\subsection{Inner Maximization: GMVFool}\label{sec:Attack}

The key to the success of VIAT in Eq.~\eqref{eq: AT formulation} is the solution to the inner maximization problem. A natural way to solve the problem is to parameterize the distribution of adversarial viewpoints with trainable parameters. The previous method ViewFool~\cite{dong2022viewfool} adopts a unimodal Gaussian distribution and performs black-box optimization based on natural evolution strategies (NES)~\cite{wierstra2014natural}. However, due to the insufficient expressiveness of the Gaussian distribution, ViewFool is unable to characterize multiple local maxima of the loss landscape w.r.t. viewpoint changes, as shown in Fig.~\ref{fig:0}. Thus, performing adversarial training with ViewFool is prone to overfitting to the specific attack and leads to poor generalization performance, as validated in the experiment. To alleviate this problem, we propose \textbf{GMVFool}, which learns a Gaussian mixture distribution of adversarial viewpoints to cover multiple local maxima of the loss landscape for more generalizable adversarial training. 

For the sake of simplicity, we omit the subscript $i$ in this subsection since the attack algorithm is the same for all objects. Specifically, we parameterize the distribution $p(\mathbf{v})$ by a mixture of $K$ Gaussian components and take the transformation of random variable approach to ensure that the support of $p(\mathbf{v})$ is contained in $[\mathbf{v}_{\min},\mathbf{v}_{\max}]$ as:
\begin{equation}\label{eq:dis}
 \mathbf{v} = \mathbf{a}\cdot \tanh(\mathbf{u})+\mathbf{b}, \; p(\mathbf{u}|\Psi) = \sum_{k=1}^{K} \mathbf{\omega}_k\mathcal{N} (\mathbf{u}|\boldsymbol{\mu}_k,\boldsymbol{\sigma}^2_k\mathbf{I}),   
\end{equation}
where $\mathbf{a}  = (\mathbf{v}_{\max}-\mathbf{v}_{\min})/2$, $\mathbf{b}  = (\mathbf{v}_{\max}+\mathbf{v}_{\min})/2$, $\Psi=\{\omega_k, \boldsymbol{\mu}_k, \boldsymbol{\sigma}_k\}^K_{k=1}$ are the parameters of the mixture Gaussian distribution with weight $\omega_k\in [0,1]$ ($\sum_{k=1}^K\mathbf{\omega }_k=1$), mean $\boldsymbol{\mu}_k\in \mathbb{R} ^{6}$ and standard deviation $\boldsymbol{\sigma }_k\in \mathbb{R} ^{6}$ of the $k$-th Gaussian component. Note that in Eq.~\eqref{eq:dis}, $\mathbf{u}$ actually follows a mixture Gaussian distribution while $\mathbf{v}$ is obtained by a transformation of $\mathbf{u}$ for proper normalization.  

Now the probability density function $p(\mathbf{u}|\Psi)$ is in the summation form, which is hard to calculate the gradients. Thus, we introduce a latent one-hot vector $\boldsymbol{\Gamma}=[\gamma_1, ..., \gamma_K]$ determining which Gaussian component the sampled viewpoint belongs to, and obeying a multinomial distribution with probability $\omega_k$, as $p(\boldsymbol{\Gamma}|\Psi)=\prod_{k=1}^{K} \mathbf{\omega}_k^{\gamma_k}$. 
With the latent variables, we represent  $p(\mathbf{u}|\Psi)$ as a multiplication form with $\boldsymbol{\Gamma}$ as $p(\mathbf{u},\boldsymbol{\Gamma}|\Psi) = \prod_{k=1}^{K} \mathbf{\omega}_k^{\gamma_k}\mathcal{N}(\mathbf{u}|\boldsymbol{\mu}_k,\boldsymbol{\sigma}^2_k\mathbf{I})^{\gamma_k}$
and $p(\mathbf{u}|\Psi)=\sum_{\Gamma}p(\mathbf{u},\boldsymbol{\Gamma}|\Psi)$, which is convenient for taking derivatives w.r.t. distribution parameters $\Psi$. 

Given the parameterized distribution $p(\mathbf{v})$ defined in Eq.~\eqref{eq:dis}, the inner maximization problem of  Eq.~\eqref{eq: AT formulation} becomes:
\begin{equation}
    \begin{split}
        \max_{\Psi}\; \mathbb{E}_{ p(\mathbf{u},\boldsymbol{\Gamma}|\Psi)} & \big[ \mathcal{L}(f_\mathbf{W}  (\mathcal{R}(\mathbf{a}\cdot \tanh(\mathbf{u})+\mathbf{b}) ) ,y) \\
        & - \lambda\cdot \log p(\mathbf{a}\cdot \tanh(\mathbf{u})+\mathbf{b})\big],
    \end{split}
    \label{eq: attack2}
\end{equation}
where the second term is the negative log density, whose expectation is the distribution's entropy (proof in Appendix). 

To solve this optimization problem, we adopt gradient-based methods to optimize the distribution parameters $\Psi$.
To back-propagate the gradients from random samples to the distribution parameters, we can adopt the low-variance reparameterization trick~\cite{blundell2015weight,kingma2013auto}. Specifically, we reparameterize $\mathbf{u}$ as 
$\mathbf{u} =  {\prod_{k=1}^{K}}\boldsymbol{\mu}^{\gamma_k}_k+{ \prod_{k=1}^{K}}\boldsymbol{\sigma}^{\gamma_k}_k\cdot\mathbf{r}$, where $\mathbf{r}\sim \mathcal{N}(\mathbf{0},\mathbf{I})$. 
With this reparameterization, the gradients of the loss function in Eq.~\eqref{eq: attack2} w.r.t. $\Psi$ can be calculated. However, similar to ViewFool, although the rendering process of NeRF is differentiable, it requires significant memory overhead to render the full image. Thus, we also resort to NES to obtain the natural gradients of the classification loss with only query access to the model. For the entropic regularizer, we directly compute its true gradient. Therefore, the gradients of the objective function in Eq.~\eqref{eq: attack2} w.r.t. $\omega_k, \boldsymbol{\mu}_k$ and $\boldsymbol{\sigma}_k$ can be derived as (proof in Appendix):
\begin{equation}\small
    \begin{split}
    \nabla_{\omega_k} = \mathbb{E}_{\mathcal{N}(\mathbf{r}|\mathbf{0},\mathbf{I})} & \left \{   \gamma_k \cdot \left [\mathcal{L}_\text{cls} \cdot \frac{1}{\omega _k} - \lambda \right ]\right \}; \\
    \nabla_{\boldsymbol{\mu}_k} = \mathbb{E}_{\mathcal{N}(\mathbf{r}|\mathbf{0},\mathbf{I})}& \left \{\gamma _k \cdot \left [\mathcal{L}_\text{cls} \cdot \frac{\boldsymbol{\sigma}_k \mathbf{r} }{\omega _k} - \lambda \!\cdot\! 2 \tanh(\boldsymbol{\mu} _k+\boldsymbol{\sigma}_k \mathbf{r}  )\right]\!\right \}\!; \\
    \nabla_{\boldsymbol{\sigma}_k} = \mathbb{E}_{\mathcal{N}(\mathbf{r}|\mathbf{0},\mathbf{I})}&\left \{\gamma _k \cdot \left [\mathcal{L}_\text{cls} \cdot \frac{\boldsymbol{\sigma_k} (\mathbf{r}^2-1) }{2\omega_k} \right.\right.\\
    &\left.\left. + \lambda \cdot \frac{(1-2\mathbf{r}\cdot \tanh(\boldsymbol{\mu} _k+\boldsymbol{\sigma}_k \mathbf{r}  )\cdot{\boldsymbol{\sigma}_k}  }{\boldsymbol{\sigma}_k} \right] \right \}; \\
    \mathcal{L}_\text{cls} = \mathcal{L}(f_\mathbf{W}(\mathcal{R}(&\mathbf{a} \cdot \tanh({\prod_{k=1}^{K}}\boldsymbol{\mu}^{\gamma_k}_k+{ \prod_{k=1}^{K}}\boldsymbol{\sigma}^{\gamma_k}_k\cdot\mathbf{r})+\mathbf{b})),y).
    \end{split}
    \label{eq: 12}
\end{equation}
In practice, we use the Monte Carlo method to approximate the expectation in gradient calculation and use iterative gradient ascent to optimize the distribution parameters of each Gaussian component. In addition, we normalize $\omega_k$ after each iteration to satisfy $ {\textstyle \sum_{k=1}^{K}} \omega_k=1$. Algorithm~\ref{alg1} outlines the overall algorithm of GMVFool.

\begin{algorithm}[htb]
	\renewcommand{\algorithmicrequire}{\textbf{Input: }}
	\renewcommand{\algorithmicensure}{\textbf{Output: }}
	\caption{GMVFool}
	\label{alg1}
	\begin{algorithmic}[1]
	    \REQUIRE Image classifier $f_\mathbf{W}$, rendering function $\mathcal{R}$, true label $y$, number of iterations $T$, number of Monte Carlo samples $q$, learning rate $\eta$,  number of Gaussian components $K$, and balance hyperparameter $\lambda$.
	    \STATE Initialize the Gaussian mixture distribution parameters of the object $\Psi^0=\{\omega^0_k , \boldsymbol{\mu}^0_k, \boldsymbol{\sigma}^0_k \}_{k=1}^K$;
	    \FOR{$t=1$ to $T$}
	    \STATE Sample $\{\mathbf{r}_j\}_{j=1}^q$ from $\mathcal{N}(\mathbf{0},\mathbf{I})$;
	    \STATE Sample $\{\boldsymbol{\Gamma}_j\}_{j=1}^q$ from the multinomial distribution with probability $\omega^t_k$;
	    \STATE Calculate $\{\mathbf{u}_j\}_{j=1}^q$;
	    \STATE Calculate $\nabla_{\Psi^t}=\{\nabla_{\omega^t_k},\nabla_{\boldsymbol{\mu}^t_k},\nabla_{\boldsymbol{\sigma}^t_k}\}$ by Eq.~\eqref{eq: 12};
	    \STATE Update the parameters:
            \STATE \quad $\Psi^{t+1} \leftarrow \Psi^{t}+\eta \cdot \nabla_{\Psi^t}$;
	    \STATE Normalize $\omega^{t+1}_k \leftarrow \omega^{t+1}_k/ {\textstyle \sum_{k=1}^{K}}\omega^{t+1}_k$;
	    \ENDFOR
		\ENSURE  Parameters of adversarial viewpoint distribution: $\Psi^T=\{\omega^T_k , \boldsymbol{\mu}^T_k, \boldsymbol{\sigma}^T_k\}_{k=1}^K$.  
	\end{algorithmic}
\end{algorithm}

\subsection{Outer Minimization}\label{sec:Defense}

In outer minimization of Eq.~\eqref{eq: AT formulation}, our goal is to minimize the loss expectation over the learned adversarial viewpoint distributions. However, there are two problems of adversarial training: inefficiency and overfitting. We next detail how we address these two problems with the \textbf{stochastic update strategy} and \textbf{distribution sharing strategy}, respectively. 

Although we introduce the efficient Instant-NGP, the inner maximization still needs many gradient steps to converge for rendering images from new viewpoints. This process is typically unacceptable for adversarial training, as each optimization step of outer minimization needs to solve the inner maximization problem for a batch of objects. To accelerate adversarial training, we propose a stochastic update strategy for the inner problem. First, we perform full inner optimization to generate adversarial viewpoint distributions for all objects given a pre-trained image classifier. At each fine-tuning epoch, we only update the distribution parameters for one randomly selected object in each category while keeping those for other objects unchanged. Note that all objects can be sufficiently optimized within multiple epochs. The rationale is that GMVFool is able to learn a sufficiently wide range of adversarial viewpoints, making the distribution effective for adversarial training over an extended period. This strategy can significantly improve efficiency and make adversarial training feasible. 

\begin{figure}[t]
  \centering
  \includegraphics[width=0.89\linewidth]{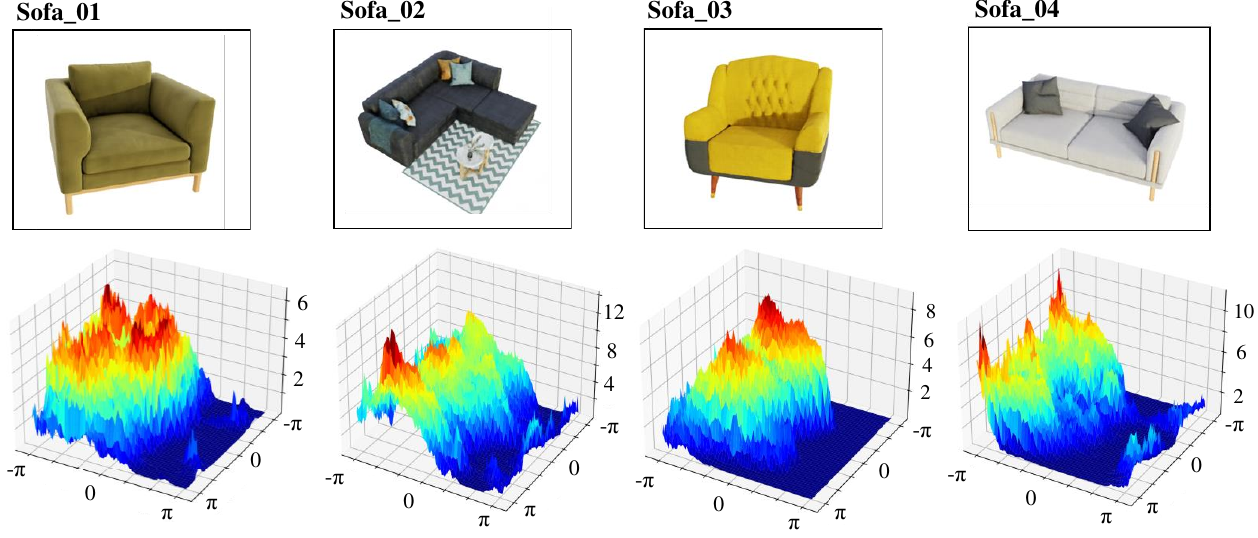}
   \caption{The adversarial viewpoint regions of objects within the same class are similar. We show the \textbf{loss landscape} w.r.t. $\psi$ and $\phi$ of four different sofas based on ResNet-50, in which we keep $[\theta, \Delta_x, \Delta_y, \Delta_z]$ as $[0, 0, 0, 0]$.}
   \vspace{-0.3cm}
   \label{fig:transferable}
\end{figure}

Besides, we find that as the training epochs increase, the learned adversarial viewpoint distributions would degenerate, \ie, in the late stage of training, the diversity of adversarial viewpoints decreases, leading to overfitting of adversarial training and inferior results.
To alleviate this problem, we propose a distribution sharing strategy, in which we share the distribution parameters of different objects within the same category. 
It is based on our
finding that the adversarial viewpoint distributions of objects within the same class are highly similar, as shown in Fig.~\ref{fig:transferable}.
For each object in training, we choose its own distribution parameters or randomly select other distribution parameters of another object  for sampling based on a probability $\pi$.

The training process of VIAT can be summarized as follows: at each fine-tuning epoch, the parameters of the adversarial viewpoint distribution are updated using GMVFool. In particular, all objects' parameters are initially optimized at the first epoch, while in each subsequent epoch, the object's parameters  within each class are randomly updated. Next, the adversarial viewpoints are sampled from a distribution based on the sharing probability. Then the corresponding adversarial examples are generated by Instant-NGP. These examples are fed to the network with the clean samples from ImageNet to calculate the cross-entropy loss. Finally, the network parameters are optimized to obtain a viewpoint-invariant model.

Moreover, VIAT boasts acceptable time costs, which can be attributed to three factors: \textbf{\underline{(1)}} It optimizes the mixture distribution of adversarial viewpoints instead of individual ones, allowing for the generation of diverse adversarial viewpoints through distribution sampling instead of multiple optimizations. \textbf{\underline{(2)}} Using the efficient Instant-NGP accelerates the optimization of adversarial viewpoint distribution and rendering of viewpoint samples. \textbf{\underline{(3)}} The stochastic optimization strategy based on distribution transferability further reduces time consumption in adversarial training.

\begin{table*}[t]

\setlength\tabcolsep{6.3pt}
\renewcommand\arraystretch{1.0}
\centering
\begin{tabular}{c|c|ccc|ccc}
\hline
\multirow{2}{*}{}        & \multirow{2}{*}{Method} & \multicolumn{3}{c|}{ResNet-50}                                                                                      & \multicolumn{3}{c}{ViT-B/16}                                                                                           \\ \cline{3-8} 
                         &                         & \multicolumn{1}{c|}{ImageNet} & \multicolumn{1}{c|}{ViewFool}       & GMVFool        & \multicolumn{1}{c|}{ImageNet} & \multicolumn{1}{c|}{ViewFool}       & GMVFool        \\ \hline \hline
Standard-trained                 & -                       & \multicolumn{1}{c|}{85.60}    & \multicolumn{1}{c|}{8.28}           & 8.98           & \multicolumn{1}{c|}{92.88}    & \multicolumn{1}{c|}{25.70}          & 29.10          \\ \hline
\multirow{2}{*}{Augmentation} & Natural                 & \multicolumn{1}{c|}{85.76}   & \multicolumn{1}{c|}{16.52}          & 19.30          & \multicolumn{1}{c|}{92.78}   & \multicolumn{1}{c|}{43.32}          & 46.48          \\
                         & Random                  & \multicolumn{1}{c|}{85.82}    & \multicolumn{1}{c|}{34.80}          & 33.52          & \multicolumn{1}{c|}{92.78}    & \multicolumn{1}{c|}{62.03}          & 67.34          \\ \hline
\multirow{2}{*}{VIAT}    & ViewFool                & \multicolumn{1}{c|}{85.66}    & \multicolumn{1}{c|}{55.12}          & 58.75          & \multicolumn{1}{c|}{92.70}   & \multicolumn{1}{c|}{79.53}          & 82.03          \\
                         & GMVFool                 & \multicolumn{1}{c|}{85.70}   & \multicolumn{1}{c|}{\textbf{59.84}} & \textbf{59.61} & \multicolumn{1}{c|}{92.56}  & \multicolumn{1}{c|}{\textbf{82.81}} & \textbf{83.13} \\ \hline
\end{tabular}
\vspace{1ex}
\caption{The \textbf{classification accuracy} (\%) from evaluation protocols with ResNet-50 and ViT-B/16, which are trained via ImageNet subset only (standard-trained), data augmentation by natural and random viewpoint images, and VIAT framework with ViewFool and GMVFool.}
\vspace{-1ex}
\label{table:defense}
\end{table*}

\begin{figure*}[tp]
  \centering
  \includegraphics[width=0.97\linewidth]{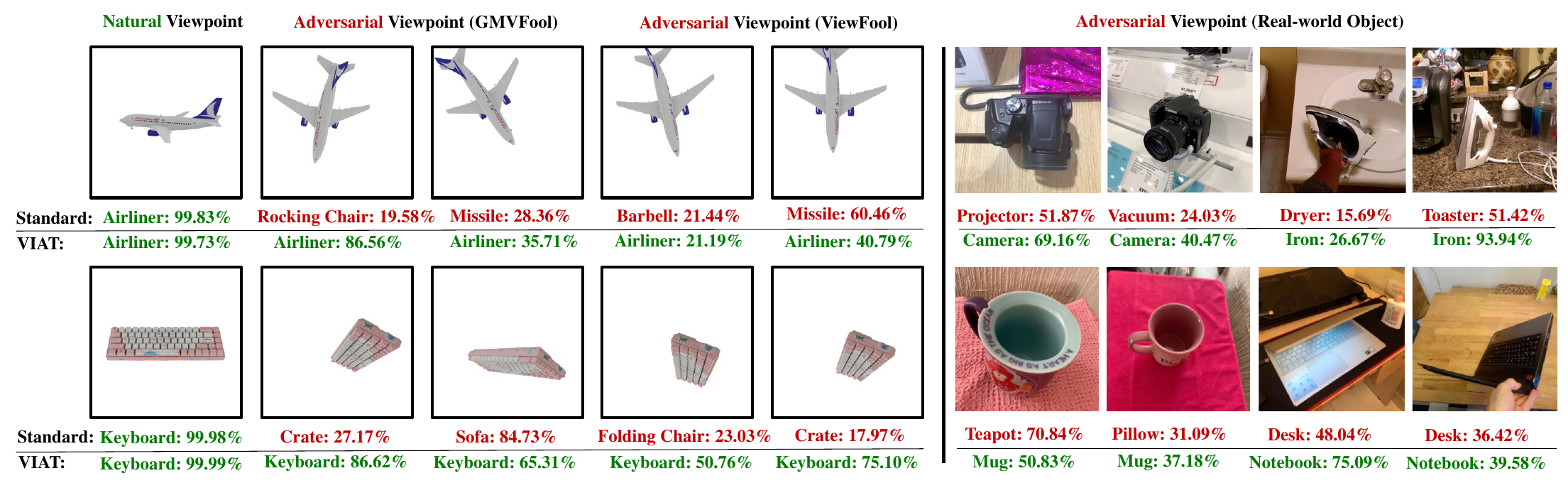}

   \caption{The \textbf{prediction} examples of Standard-trained and VIAT-trained ResNet-50 under natural and adversarial viewpoint images. Green and red text represent correct and incorrect predictions, respectively, and the corresponding number is the confidence value.}\vspace{-1ex}
   \label{fig:visualization}
   \vspace{-0.3cm}
\end{figure*}

\section{Experiments}
\vspace{-0.1cm}
\subsection{Performance of VIAT}\label{sec:defense exp}
\vspace{-0.1cm}
\textbf{Experimental settings.} \textbf{(A) Datasets:} To address the lack of multi-view images in ImageNet, we construct IM3D: a dataset composed of 1K typical synthetic 3D objects from 100 ImageNet categories, each category containing 10 objects. We acquire the multi-view images on the upper hemisphere with the corresponding camera poses for each object, and then we learn the NeRF representation using Instant-NGP. Objects of each type are divided into a training set and a validation set with a ratio of 9:1, which are utilized for adversarial training and validation of viewpoint invariance. \textbf{(B) Model:} Two classifiers are considered to perform experiments, including the CNN-based ResNet-50~\cite{he2016deep} and the Transformer-based ViT-B/16~\cite{dosovitskiy2020image}. We train the classifiers on the subset of ImageNet, which corresponds to our 3D synthetic object’s category, achieving 85.60\%, 92.88\% Top-1 accuracy on the ImageNet subset.
\textbf{(C) VIAT Setting:} Following~\cite{dong2022viewfool}, we initialize the camera at $[0,4,0]$, the range of rotation parameters are set as $\psi \in [-180^{\circ}, 180^{\circ}]$, $\theta \in [-30^{\circ}, 30^{\circ}]$, $\phi \in [20^{\circ}, 160^{\circ}]$, the range of translation parameters are set as $\Delta_x \in [-0.5, 0.5]$, $\Delta_y \in [-1, 1]$, $\Delta_z \in [-0.5, 0.5]$, and the balance hyperparameter $\lambda=0.01$. Based on the results of the ablation studies, we set the components number $K$=15, and distribution sharing probability $\pi$=0.5. For the inner maximization step, we approximate the gradients in Eq.~\eqref{eq: 12} with $q$=100 MC samples and use the Adam~\cite{kingma2014adam} optimizer to update $\Psi$ for 50 iterations in the first epoch, then iterate 10 times under the previous $\Psi$ for subsequent epochs. After obtaining the model trained on the ImageNet subset, we continue to train the model for 60 epochs with the adversarial viewpoints and ImageNet clean samples, with a ratio of 1:32.

\textbf{Evaluation metrics.}
To fully explore the viewpoint invariance of models, we use Top-1 accuracy as the evaluation metric and set up four evaluation protocols: (a) ImageNet: the accuracy is calculated under the validation set of ImageNet. (b) ViewFool: the accuracy is calculated under renderings of adversarial viewpoints generated by ViewFool. (c) GMVFool: the accuracy is calculated under renderings of adversarial viewpoints generated by our GMVFool.

\begin{figure}[t]
  \centering
  \includegraphics[width=0.94\linewidth]{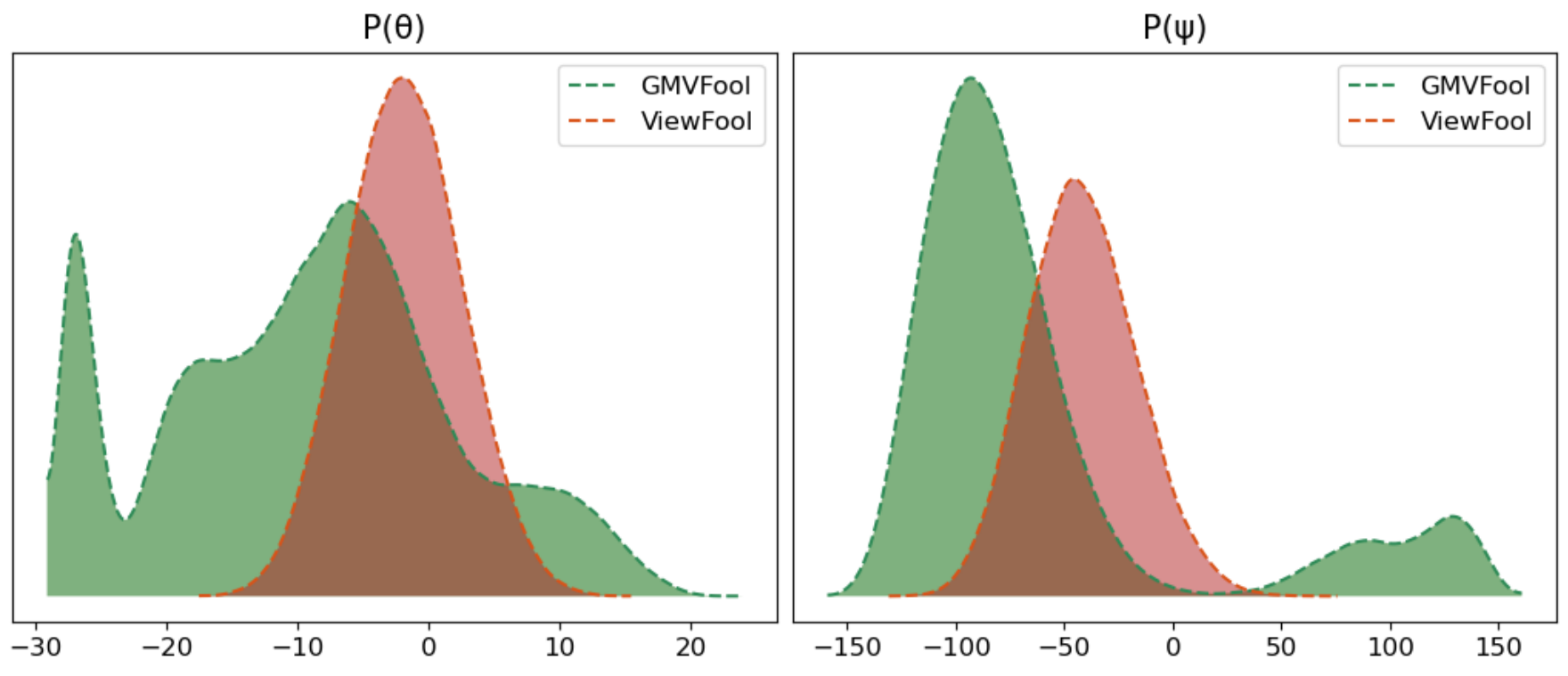} 
   \vspace{-1ex}\caption{The \textbf{probability density curves} of the adversarial distribution under viewpoint parameters $\theta$ and $\psi$, which are optimized by GMVFool ($K=5$) and ViewFool, respectively.} \vspace{-3ex}
   \label{fig:distribution}
\end{figure}

\textbf{Experimental results.}
The experimental results are shown in Table~\ref{table:defense}. We compare VIAT with three baselines: (a) Data augmentation with the most common viewpoint renderings from training objects’ natural states. For this, we define a range of views frequently appearing in ImageNet for each class (\eg hotdogs are usually in the top view). (b) Data augmentation with random viewpoint rendering of objects in the training set. (c) VIAT uses ViewFool as the inner maximization method. To be fair, we also use Instant-NGP for accelerating ViewFool in the adversarial training. From the table, we can draw the following conclusions: 

\textbf{\underline{(1)}}
VIAT significantly improves the viewpoint invariance of the model. Under the adversarial viewpoint generated by GMVFool and ViewFool,  the accuracy of ResNet-50 is improved by 50.63\% and 51.56\%, while that of ViT-B/16 is improved by 54.03\% and 57.11\%, respectively, compared to the standard-trained model.

\textbf{\underline{(2)}}
Adopting GMVFool as VIAT’s inner maximization method results in higher accuracy than adopting ViewFool under the adversarial viewpoint images. The smaller accuracy gap between the two attack methods means that VIAT+GMVFool can better generalize to different viewpoint attacks. We think it benefits from the Gaussian mixture modelling of GMVFool which can generate diverse adversarial viewpoints to be learned by the network.

\textbf{\underline{(3)}}
Data augmentation methods using natural and random viewpoint images have great limitations in improving the model's performance under adversarial viewpoints.

\textbf{\underline{(4)}}
ViT-B/16 is better than ResNet-50 in resisting adversarial viewpoint attacks. This phenomenon may benefit from its transformer structure, which is also confirmed by the benchmark results of ImageNet-V+ in Sec.~\ref{sec:imagenet-v+}. 

\textbf{Visualization.} Fig.~\ref{fig:visualization} shows the visualization results of the natural and adversarial viewpoints rendering of objects, as well as the output and confidence of the standard-trained and VIAT-trained ResNet-50. The results demonstrate that VIAT-trained model can still predict the correct labels when facing the adversarial viewpoint. Additional examples will be presented in the Appendix.
\vspace{-1ex}
\subsection{Additional Results and Ablation Studies}
\begin{figure}[htbp]
	\centering
	\begin{minipage}{0.46\linewidth}
		\centering
		\includegraphics[width=0.99\linewidth]{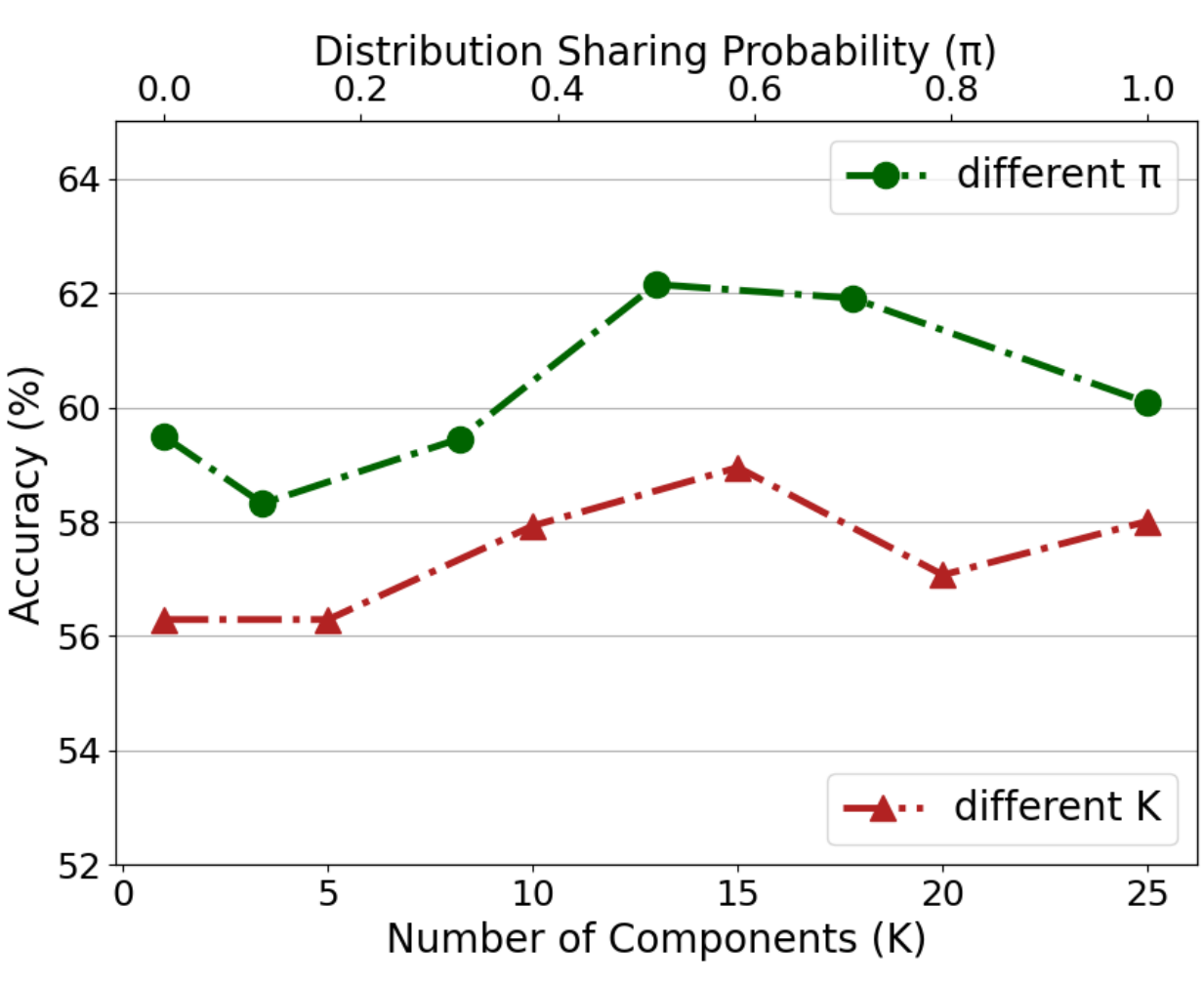}
		\caption{\footnotesize The \textbf{accuracy} (\%) of VIAT-trained ResNet-50 against adversarial viewpoints, using various sets of $K$ and $\pi$.}
		\label{fig:ablation}
	\end{minipage} \hspace{0.3cm}
	\begin{minipage}{0.46\linewidth}
		\centering
		\includegraphics[width=0.99\linewidth]{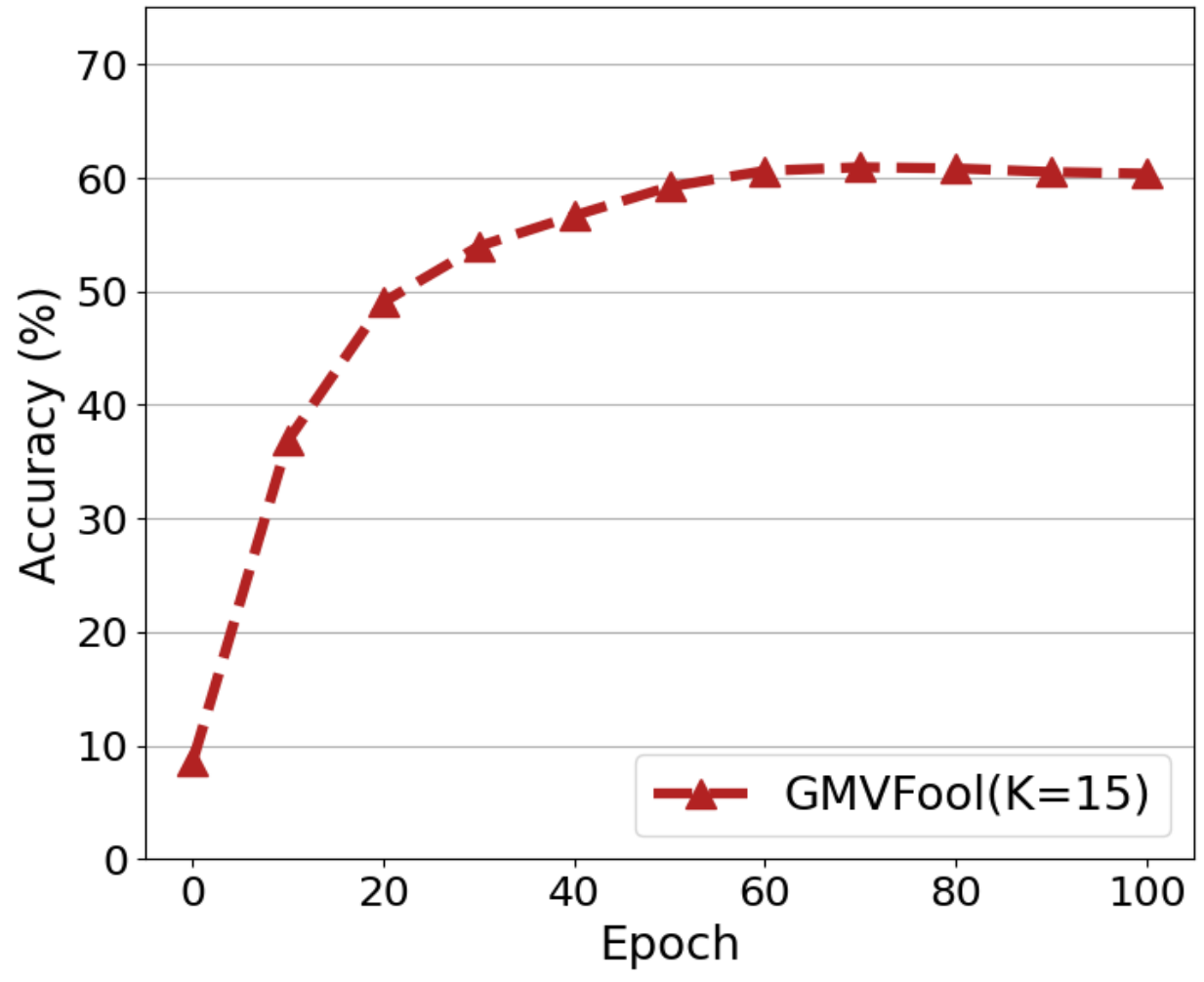}
		\caption{\footnotesize The \textbf{accuracy} (\%) of VIAT-trained ResNet-50 against GMVFool attack with different training iterations.}
		\label{fig:convergence}
	\end{minipage}
 \vspace{-0.3cm}
\end{figure}

\vspace{-0.1cm}
\label{sec:ablation}
\textbf{The effects of $K$ and $\pi$.} We further conduct ablation experiments to investigate the impact of the number of Gaussian components ($K$) and distribution sharing probability ($\pi$). Fig.~\ref{fig:ablation} presents the classification accuracy of the model against GMVFool attacks after VIAT training with different settings. We observe a positive correlation between the model's ability to resist viewpoint attacks and the number of components used by VIAT. Additionally, a suitable sharing probability benefits the model in achieving better viewpoint invariance. However, a high component number and sharing probability will lead to the opposite situation.




\textbf{Convergence discussion.} As a distribution-based adversarial training framework, the convergence of VIAT is guaranteed in theory~\cite{dong2020adversarial}. Furthermore, We study the convergence of VIAT with learning rate 0.001, $K=15$, $\pi=0.5$. The accuracy under adversarial viewpoints generated by GMVFool is presented in Fig.~\ref{fig:convergence}, indicating that VIAT can converge well under experimental setting.

\begin{figure*}[tp]
  \centering
  \includegraphics[width=0.92\linewidth]{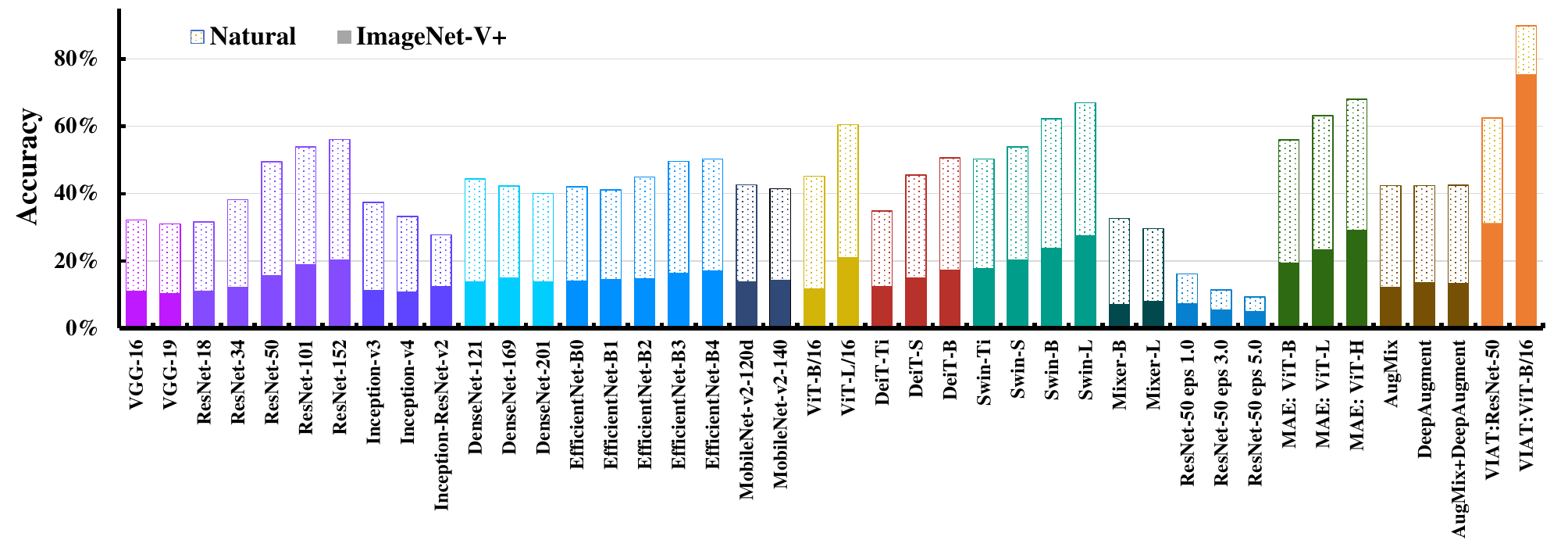}\vspace{-2ex}
   \caption{The \textbf{accuracy} of different classifiers on natural viewpoint and on ImageNet-V+.}\vspace{-0.5cm}
   \label{fig:benchmark}
   \vspace{0.2cm}
\end{figure*} 


\textbf{The superiority of GMVFool.} Contributed by the mixture distribution design, we can control $K$ to balance viewpoint diversity and attack performance in different tasks. Specifically, for adversarial training, GMVFool with larger $K$ is used to improve the adversarial viewpoint diversity, which is crucial for achieving better robustness via adversarial training. As depicted in Fig.~\ref{fig:distribution}, GMVFool ($K\!\!=\!\!5$) captures more comprehensive and diverse adversarial viewpoints than ViewFool. Table~\ref{table:attack} quantitatively confirms this observation through the entropy $\mathcal{H}(p^*(\mathbf{v}))$. For viewpoint attacks, GMVFool can be more effective with a smaller $K$. Table~\ref{table:attack} compares the attack performance of GMVFool ($K\!\!=\!\!1$, $3$, and $5$, respectively) with previous methods. It shows that GMVFool ($K\!\!=\!\!1$) performs best on the \emph{optimal distribution of adversarial viewpoints} $p^*(\mathbf{v})$ and GMVFool ($K\!\!=\!\!3$ and $5$) maintains good performance under $p^*(\mathbf{v})$. 




\begin{table*}[tbp]
\small
\setlength\tabcolsep{2.8pt}
\renewcommand\arraystretch{1.0}
\centering
\begin{tabular}{c|cc|cc|cc|cc}
\hline
\multirow{2}{*}{Method} & \multicolumn{2}{c|}{ResNet-50~\cite{he2016deep}}                                                       & \multicolumn{2}{c|}{EN-B0~\cite{tan2019efficientnet}}                                                           & \multicolumn{2}{c|}{DeiT-B~\cite{touvron2021training}}                                                          & \multicolumn{2}{c}{Swin-B~\cite{liu2021swin}}                                                           \\ \cline{2-9} 
                        & \multicolumn{1}{c|}{$\mathcal{R}(p^*(\mathbf{v}))\uparrow$} & $\mathcal{H}(p^*(\mathbf{v}))\uparrow$ & \multicolumn{1}{c|}{$\mathcal{R}(p^*(\mathbf{v}))\uparrow$} & $\mathcal{H}(p^*(\mathbf{v}))\uparrow$ & \multicolumn{1}{c|}{$\mathcal{R}(p^*(\mathbf{v}))\uparrow$} & $\mathcal{H}(p^*(\mathbf{v}))\uparrow$ & \multicolumn{1}{c|}{$\mathcal{R}(p^*(\mathbf{v}))\uparrow$} & $\mathcal{H}(p^*(\mathbf{v}))\uparrow$ \\ \hline\hline
Random Search           & \multicolumn{1}{c|}{64.12}                          & -                              & \multicolumn{1}{c|}{79.18}                          & -                              & \multicolumn{1}{c|}{32.25}                          & -                              & \multicolumn{1}{c|}{19.88}                          & -                              \\
ViewFool                & \multicolumn{1}{c|}{91.50}                          & -10.14                         & \multicolumn{1}{c|}{95.64}                          & -10.38                         & \multicolumn{1}{c|}{80.37}                          & -10.41                         & \multicolumn{1}{c|}{73.85}                          & -10.71                         \\ \hline
GMVFool (K=1)            & \multicolumn{1}{c|}{\textbf{92.13}}                 & -10.17                         & \multicolumn{1}{c|}{\textbf{95.78}}                 & -10.38                         & \multicolumn{1}{c|}{\textbf{80.61}}                 & -10.44                         & \multicolumn{1}{c|}{\textbf{74.02}}                 & -10.62                         \\
GMVFool (K=3)            & \multicolumn{1}{c|}{89.20}                          & -3.75                          & \multicolumn{1}{c|}{\textbf{95.78}}                          & -3.94                          & \multicolumn{1}{c|}{79.65}                          & -3.92                          & \multicolumn{1}{c|}{72.48}                          & -4.05                          \\
GMVFool (K=5)            & \multicolumn{1}{c|}{91.56}                          & \textbf{-0.69}                 & \multicolumn{1}{c|}{95.62}                          & \textbf{-0.85}                 & \multicolumn{1}{c|}{78.91}                          & \textbf{-1.00}                 & \multicolumn{1}{c|}{70.16}                          & \textbf{-1.07}                 \\ \hline
\end{tabular}
\vspace{0.1cm}
\caption{The \textbf{attack success rate} (\%) and the \textbf{entropy} of methods against various classifiers. $\mathcal{R}(\cdot)$ denotes the rendering process.}
\label{table:attack}
\vspace{-0.5cm}
\end{table*}

\vspace{-0.1cm}
\subsection{Evaluation on Other Datasets}
\vspace{-0.1cm}
\textbf{Performance on real-world adversarial viewpoints.} We conduct evaluation experiments on objectron~\cite{ahmadyan2021objectron}, which contains object-centric videos in the wild.. The accuracy of standard-trained and VIAT-trained ResNet-50 is presented in Table~\ref{table:objectron}, demonstrating that VIAT enhances the model's performance in unnatural viewpoints. Some cases are shown in Fig.~\ref{fig:visualization}. Furthermore, we conduct experiments on other OOD datasets that contain viewpoint perturbations, the results in Table~\ref{table:ood} demonstrate the VIAT-trained model's ability to resist different natural perturbations.

\vspace{-0.3cm}
\begin{table}[htb]
\small
\setlength\tabcolsep{5.5pt}
\renewcommand\arraystretch{1.0}
\centering
\begin{tabular}{cl|c|c|c|c|c}
\hline
\multicolumn{2}{c|}{}   & \emph{shoe}           &  \emph{camera}         &  \emph{mug}            &  \emph{chair}          &  \emph{computer}       \\ \hline \hline
\multicolumn{2}{c|}{Standard} & 76.28          & 81.42          & 97.67          & 38.82          & 60.19          \\
\multicolumn{2}{c|}{VIAT (Ours)}     & \textbf{76.73} & \textbf{86.93} & \textbf{99.07} & \textbf{39.75} & \textbf{63.40} \\ \hline
\end{tabular}
\vspace{0.1cm}
\caption{The \textbf{accuracy} (\%) of standard-trained and VIAT-trained ResNet-50 in the real-world images from the objectron dataset.}
\label{table:objectron}
\vspace{-0.3cm}
\end{table}
\textbf{Performance on multi-view datasets.} We conduct evaluation experiments on other multi-view datasets, for which we obtain the categories that overlap with our training categories. Table~\ref{table:ood} presents the accuracy of the model under various viewpoints, indicating the improved robustness of the VIAT-trained model against viewpoint transformations.


\begin{table}[htb]
\small
\setlength\tabcolsep{3.5pt}
\renewcommand\arraystretch{1.0}
\centering
\begin{tabular}{c|l|ccc}
\hline
                                                                                & Dataset    & Standard &Aug & VIAT (Ours) \\ \hline \hline
\multirow{3}{*}{\begin{tabular}[c]{@{}c@{}}OOD\\ Datasets\end{tabular}}         & ObjectNet~\cite{barbu2019objectnet}  & 35.92               & 36.02                      & \textbf{37.39}  \\  
                                                                                & ImageNet-A~\cite{hendrycks2021natural} & 19.03                &  18.65                     & \textbf{20.32}  \\ 
                                                                                & ImageNet-R~\cite{hendrycks2021many} & 45.06               & 45.09                      & \textbf{46.51}  \\ 
                                                                                & ImageNet-V~\cite{dong2022viewfool} & 28.15               & 32.83                      & \textbf{38.96}  \\ \hline
\multirow{3}{*}{\begin{tabular}[c]{@{}c@{}}Multi-view \\ Datasets\end{tabular}} & MIRO~\cite{kanezaki2018rotationnet}       & 57.41                    & 58.78                      &  \textbf{65.86}               \\ 
                                                                                &
                                                                                OOWL~\cite{ho2019catastrophic}        & 51.41                    & 51.24                      & \textbf{52.13}               \\ 
                                                                                & CO3D~\cite{reizenstein2021common}       & 64.68                    &  64.90                     &  \textbf{66.04}               \\ \hline
\end{tabular}
\vspace{1px}
\caption{The \textbf{accuracy} (\%) of standard-trained, Augmentation with random renderings (Aug) and VIAT-trained ResNet-50 in various OOD and multi-view datasets.}
\vspace{-0.4cm}
\label{table:ood}
\end{table}




\subsection{ImageNet-V+ Benchmark}
\vspace{-0.1cm}
\label{sec:imagenet-v+}

We utilize GMVFool to construct a larger benchmark dataset, ImageNet-V+, to evaluate the viewpoint robustness of visual recognition models. It comprises 100K adversarial viewpoint images of 1K synthetic objects belonging to the 100 ImageNet classes. 
The details and visualizations will be included in the Appendix. We adopt ImageNet-V+ to evaluate 40 different models pre-trained on ImageNet, including models with differen\textbf{}t structures (the CNN-based VGG~\cite{simonyan2014very}, ResNet~\cite{he2016deep}, Inception~\cite{szegedy2016rethinking,szegedy2017inception}, DenseNet~\cite{huang2017densely}, EfficientNet~\cite{tan2019efficientnet}, MobileNet-v2~\cite{sandler2018mobilenetv2}, the transformer-based: ViT~\cite{dosovitskiy2020image}, DeiT~\cite{touvron2021training}, Swin Transformer~\cite{liu2021swin}, and the MLP Mixer~\cite{tolstikhin2021mlp}), different training paradigms (adversarial training~\cite{salman2020adversarially} and mask-autoencoder~\cite{he2022masked}), different augmentation methods (AugMix~\cite{hendrycks2019augmix}, DeepAugment~\cite{hendrycks2021many}). For comparison, we also evaluate the model trained with VIAT. 

Fig.~\ref{fig:benchmark} illustrates the accuracy of various models on natural viewpoint images and ImageNet-V+. When exposed to adversarial viewpoints, the accuracy of all models decreases significantly. We observe that the model's performance with the same architectures is positively related to its size, 
with transformer-based models outperforming CNN-based models. Among them, MAE with ViT-H performs best in ImageNet-V+, achieving 29.37\% accuracy. Models using data augmentation and adversarial training, which is robust to adversarial examples and image corruption in previous work, perform poorly from the adversarial viewpoint. Finally, ViT-B/16 trained with VIAT outperform all models using standard training, achieving an accuracy of 75.49\%.


\section{Conclusion}
This paper proposed the VIAT framework to obtain viewpoint invariance for visual recognition via adversarial training and contributed GMVFool, an efficient method for generating diverse adversarial viewpoints. We also provided a new multi-view dataset---IM3D and conducted extensive experiments to verify the effectiveness of VIAT in enhancing viewpoint invariance. Moreover, we introduced ImageNet-V+, a large viewpoint OOD benchmark including 100K adversarial viewpoint images of 1K synthetic objects, and provided the accuracy on various models.

{\small
\bibliographystyle{ieee_fullname}
\bibliography{egbib}
}

\end{document}